\documentclass[10pt,twocolumn]{article}

\usepackage[letterpaper,margin=0.72in,columnsep=0.24in]{geometry}
\usepackage{amsmath,amssymb,booktabs,array,multirow}
\usepackage{graphicx}
\usepackage{microtype}
\usepackage{enumitem}
\usepackage{xcolor}
\usepackage{url}
\usepackage[hidelinks]{hyperref}
\usepackage{authblk}
\usepackage{caption}
\usepackage{dblfloatfix}
\usepackage{float}
\usepackage{placeins}
\usepackage{titlesec}
\usepackage{fancyhdr}
\usepackage{etoolbox}

\definecolor{Navy}{HTML}{1F3B5C}
\definecolor{MidGray}{HTML}{5F6670}
\setlist[itemize]{leftmargin=*,nosep}
\setlist[enumerate]{leftmargin=*,nosep}
\titleformat{\section}{\large\bfseries\color{Navy}}{\thesection}{0.5em}{}
\titleformat{\subsection}{\normalsize\bfseries}{\thesubsection}{0.45em}{}
\titleformat{\subsubsection}{\normalsize\itshape}{\thesubsubsection}{0.45em}{}
\title{\vspace{-0.7em}\textbf{Building Telecom Language Models with RSOD:}\\[-0.15em]
\textbf{Domain Data, Specialist Post-Training, Memory-Conscious Distillation, and Networking Evaluation}}

\author[1]{Anil Srirangapatna Nagesh}
\author[1]{Nitin Gupta}
\author[1]{Anuj Taneja}
\author[1]{Vivek Singh}
\author[1]{Shane Eleniak}
\author[1]{Theo Zhao}
\author[1]{Moulidharan N}
\author[1]{Shashank Keshetty}
\affil[1]{Calix AI Labs}
\date{}

\begin{document}
\maketitle
\vspace{-0.8em}
\begin{abstract}
Telecom language models must combine broad networking knowledge with specialist behavior across routing, broadband services, access technologies, and operational diagnostics while remaining practical to deploy as a single model. We present an end-to-end framework that builds source-grounded synthetic training data from structured networking taxonomies and expert graphs, materializes 212,559 routed training records across fourteen networking routes, trains six frozen specialist LoRA teachers over one shared continued-pretrained backbone, and consolidates their behavior into one deployable student adapter using Routed Specialist On-Policy Distillation (RSOD). In RSOD, the committed student generates each trajectory, the route-assigned specialist teacher-forces the exact native token sequence, compact teacher hidden states are reconstructed through the shared frozen output head, and only the student LoRA is updated. A controlled reconstruction validation shows mean absolute logit error of $7.26\times10^{-6}$ with 100\% top-1 agreement between native and reconstructed teacher logits. On four RTX 6000 Ada GPUs, the completed 2,000-step run produced 16,000 trajectories with zero observed student-version or token-ID mismatches. Under the primary route-conditioned evaluation over 2,850 held-out questions, the raw Meta-Llama-3.1-8B-Instruct model scores 12.8661/25 micro (13.0172/25 macro), the combined routed six-teacher system scores 21.7641/25 micro, and the consolidated RSOD student, released as Llama-CalixCLM v1, reaches 21.7112/25 micro (21.8006/25 macro) on the same five-dimension LLM-judge rubric. Under a separate neutral deployment-like prompt, the same student reaches 20.0028/25 micro (20.0778/25 macro). The consolidated student remains within 0.0529 micro-score points of the combined routed-teacher system while retaining a numerically validated and auditable train--serve execution path.
\end{abstract}

\section{Introduction}
Telecom language models must combine broad networking knowledge with operationally specific behavior across routing, transport, broadband services, access technologies, command-line diagnosis, and troubleshooting. Specialist adaptation can sharpen domain behavior, but deploying multiple specialist endpoints complicates model management and serving. Our framework keeps one shared backbone, trains lightweight specialist LoRA adapters, and then consolidates their behavior into one student adapter for deployment.

Parameter-efficient adaptation keeps the backbone frozen while learning low-rank updates \cite{hu2022lora}. We extend that foundation into a complete telecom post-training workflow spanning structured domain data, continued pretraining (CPT), specialist adaptation, routed on-policy distillation, auditable train--serve synchronization, and networking-specific evaluation.

\begin{figure*}[t]
    \centering
    \includegraphics[width=0.98\textwidth]{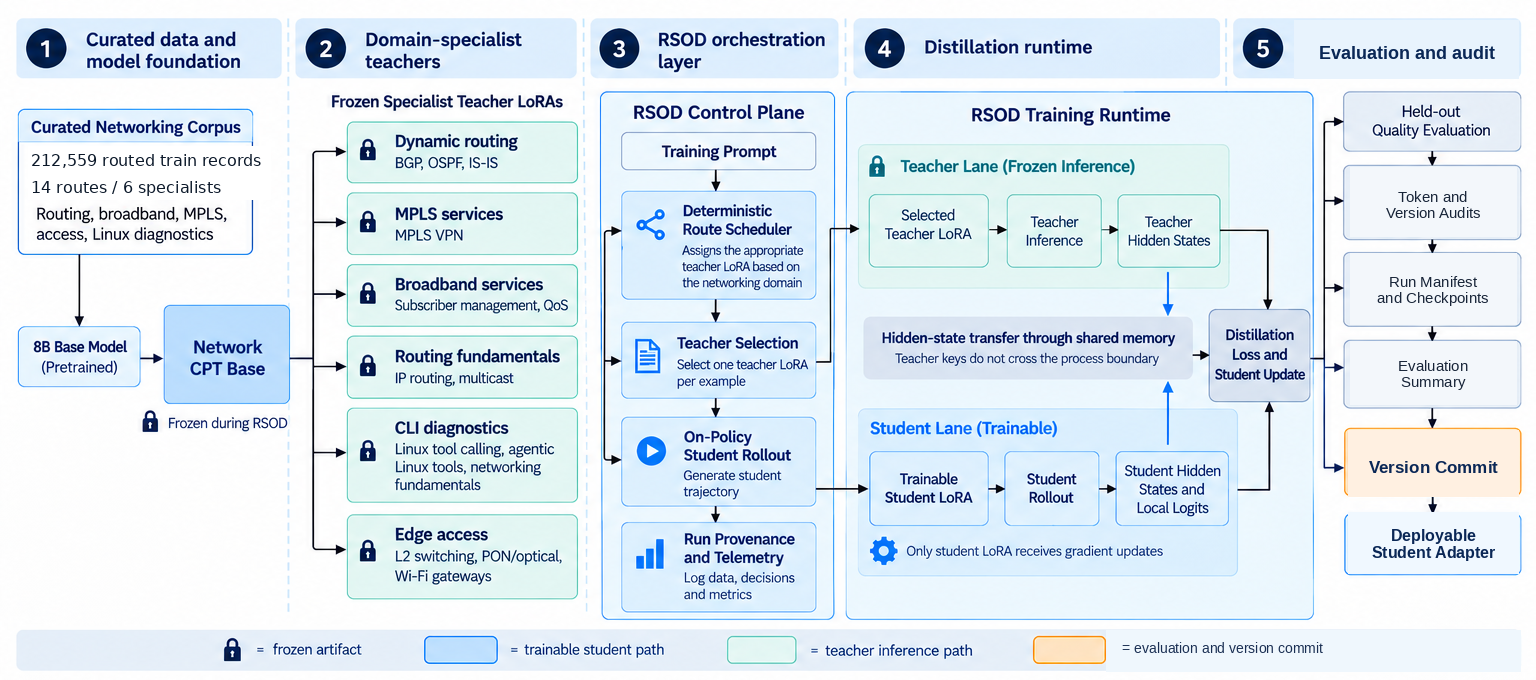}
    \caption{End-to-end framework. Structured networking data supports one shared CPT backbone and six frozen specialist LoRAs. RSOD routes each training trajectory to one specialist, trains a single student adapter, records run provenance and audits, and produces one deployable student.}
    \label{fig:framework}
\end{figure*}

Figure~\ref{fig:framework} summarizes the workflow. Our research question is: \emph{Can multiple telecom-specialized LoRA teachers be consolidated into one deployable adapter using student-on-policy distillation while retaining route-conditioned specialist quality and exact token/version consistency?} In the evaluated run, the raw model scores 12.8661/25 micro, the combined routed six-teacher system scores 21.7641/25 micro, and the consolidated student reaches 21.7112/25 micro under the primary route-conditioned protocol; all 16,000 recorded version checks and token-ID checks match.

Our contributions are:
\begin{itemize}
    \item \textbf{An end-to-end telecom specialization framework} connecting structured knowledge-grounded data generation and 212,559 routed training records to six specialist groups, fourteen canonical routes, specialist LoRA post-training, consolidation, and evaluation.
    \item \textbf{RSOD with exact routed supervision}: the committed student generates the trajectory, deterministic route ownership selects one frozen specialist, and the teacher scores the student's exact native token IDs before forward-KL optimization of only the student LoRA.
    \item \textbf{A memory-conscious execution architecture} that shares one tensor-parallel teacher backbone across specialist LoRAs, transports compact teacher hidden states for local logit reconstruction, and maintains gradients and optimizer state only for the student LoRA.
    \item \textbf{A fourteen-route telecom networking evaluation protocol} with fixed route-specific system prompts and a common five-dimension scoring framework applied consistently across compared models.
    \item \textbf{An auditable RSOD execution protocol and evidence layer} combining version-consistent rollouts, exact-token specialist supervision, compact teacher-state reconstruction, machine-verifiable token/version provenance, numerical fidelity validation, and matched systems/quality measurements.
\end{itemize}

RSOD builds on established LoRA, on-policy distillation, multi-teacher selection, and shared multi-adapter serving. Its contribution is the integrated configuration of fixed route-to-specialist assignment, student-generated trajectories, exact-token teacher forcing, shared-head signal reconstruction, version-acknowledged adapter commits, and telecom-specific data/evaluation mappings. Table~\ref{tab:positioning} summarizes this positioning.

\begin{table}[t]
\centering
\caption{Abbreviations used in the paper.}
\label{tab:abbrev}
\scriptsize
\setlength{\tabcolsep}{3pt}
\begin{tabular}{lp{0.60\columnwidth}}
\toprule
Term & Meaning \\
\midrule
RSOD & Routed Specialist On-Policy Distillation \\
CPT & Continued Pretraining \\
SFT & Supervised Fine-Tuning \\
KD & Knowledge Distillation \\
KL & Kullback--Leibler divergence \\
LoRA & Low-Rank Adaptation \\
PEFT & Parameter-Efficient Fine-Tuning \\
DDP & Distributed Data Parallel \\
TP / TP2 & Tensor Parallelism / two-way tensor parallelism \\
BF16 & Brain Floating Point 16-bit \\
VRAM & Video RAM \\
BNG & Broadband Network Gateway \\
MPLS & Multiprotocol Label Switching \\
PON & Passive Optical Network \\
QoS & Quality of Service \\
BGP & Border Gateway Protocol \\
OSPF & Open Shortest Path First \\
IS-IS & Intermediate System to Intermediate System \\
\bottomrule
\end{tabular}
\end{table}

\begin{table*}[t]
\centering
\caption{RSOD combines established building blocks into an audited telecom-specialist consolidation workflow.}
\label{tab:positioning}
\small
\begin{tabular}{p{0.16\textwidth}p{0.30\textwidth}p{0.46\textwidth}}
\toprule
Dimension & Established building block & RSOD configuration \\
\midrule
Trajectories & Student-on-policy distillation & Route-conditioned student trajectories generated by the committed student \\
Teacher choice & Multi-teacher selection or aggregation & Deterministic route-to-specialist assignment; one teacher per trajectory \\
Token alignment & Teacher scoring of generated output & Exact native token-ID teacher forcing on the student's sequence \\
Signal transport & Logits or local teacher execution & Teacher hidden-state transport with local shared LM-head logit reconstruction \\
Model serving & Shared multi-adapter serving & One shared teacher backbone plus version-acknowledged student adapter commits \\
Validation & Aggregate training metrics & Direct token-identity and serving-version audits over the run \\
\bottomrule
\end{tabular}
\end{table*}

\section{Related Work}
\textbf{Knowledge and on-policy distillation.} Knowledge distillation trains a student to match a teacher distribution using soft targets \cite{hinton2015distill}. Generalized Knowledge Distillation (GKD) addresses autoregressive train--test mismatch by training on student-generated trajectories \cite{agarwal2023gkd}. Nemotron-Cascade~2 applies multi-domain on-policy distillation from strong intermediate teachers \cite{yang2026nemotroncascade2}, while Ma et al. study multi-teacher on-policy distillation for capability integration \cite{ma2026mopd}. RSOD specializes this design space around deterministic telecom route ownership, exact-token teacher forcing, compact teacher-signal transport, and version-consistent train--serve synchronization.

\textbf{Specialist consolidation and model merging.} Multi-teacher methods study example-wise teacher selection and aggregation \cite{yuan2020multiteacher}. Parameter-space methods such as TIES-Merging combine specialized models directly \cite{yadav2023ties}. RSOD instead performs behavioral consolidation: specialist adapters remain frozen and separate while one student learns from the assigned teacher on its own trajectories.

\textbf{Parameter-efficient and multi-adapter serving.} LoRA provides frozen-backbone low-rank adaptation \cite{hu2022lora}. Punica and S-LoRA establish shared-backbone serving for many adapters \cite{chen2024punica,sheng2024slora}, and SGLang provides an efficient inference runtime \cite{zheng2024sglang}. RSOD uses shared multi-adapter serving inside a synchronized training loop in which successive rollouts use the newly committed student version.

\textbf{Telecom models and evaluation.} TeleQnA, TelecomGPT, TeleMath, Tele-LLMs, NETLLMBENCH, and the GSMA Open-Telco benchmark effort motivate telecom-specific evaluation beyond general-purpose language-model scores \cite{maatouk2023teleqna,zou2024telecomgpt,colle2025telemath,maatouk2024telellms,aykurt2024netllmbench,gsma2025opentelco}. Our evaluation applies the same route-specific prompt and five-dimension scoring protocol to the raw model, the routed CPT+specialist teacher system, and the consolidated student.

\section{Domain Data and Specialist Post-Training}
\subsection{Knowledge-Grounded Corpus Construction}
Training data are produced through a structured knowledge-generation pipeline rather than assembled as a flat question--answer collection. For each networking domain, a protocol taxonomy defines the intended knowledge space and is expanded into an expert graph containing concepts, protocol fields, semantic relationships, workflows, operational behavior, failure modes, and diagnostic evidence. Graph nodes, semantic edges, and higher-order subgraphs are converted into generation units. A planning stage translates each unit into explicit learning obligations and CPT/SFT generation intents, after which a teacher model realizes the corresponding natural-language records.

The underlying technical knowledge is source-derived, whereas the final training records are synthetic realizations. Depending on the domain, the knowledge-building stage draws from standards and RFCs, protocol registries, architecture and specification documents, vendor-neutral technical documentation, implementation evidence, operational troubleshooting knowledge, and selected vendor documentation when implementation behavior is not fully specified by standards. Synthetic generation is therefore used for linguistic realization rather than unconstrained knowledge creation: CPT passages, explanations, questions, scenarios, comparisons, diagnostic exercises, and misconception-correction examples are generated from the structured representation.

Validation is applied throughout the pipeline. Taxonomy and graph stages are checked for structural coverage, atomicity, missing concepts, relationships, and weakly grounded nodes; planning checks verify that learning obligations map to CPT/SFT intents; generated corpora are checked for missing or extra planned records, empty outputs, identifier duplication, exact duplication, repeated textual patterns, planner/meta-instruction leakage, placeholder content, length distributions, and diversity. Diagnostic evaluations used during pipeline development could inform taxonomy and graph refinement. The final 2,850-question held-out benchmark reported in this paper was frozen separately and was not used to drive taxonomy, expert-graph, or training-data revisions.

\subsection{Routed Training and Held-Out Inventory}
The executed experiment materializes 212,559 routed training records across fourteen canonical routes and six specialist groups. The same route identity is carried from data preparation into specialist ownership, deterministic teacher selection, and route-specific evaluation. Each routed JSONL artifact is recorded with a SHA-256 identity in the preparation manifest. Training and held-out artifacts are physically separate: CPT and specialist SFT consume routed train files, while the evaluation inventory contains 2,850 held-out questions in independently hashed files.

\begin{table}[t]
\centering
\caption{Routed training and held-out inventory used by the final six-specialist configuration.}
\label{tab:specialists}
\small
\begin{tabular}{lrr}
\toprule
Specialist & Train records & Heldout \\
\midrule
Dynamic Routing & 45,597 & 570 \\
MPLS Services & 26,439 & 190 \\
Broadband Services & 29,360 & 570 \\
Routing Fundamentals & 52,083 & 380 \\
CLI Diagnostics & 20,990 & 570 \\
Edge Access & 38,090 & 570 \\
\midrule
Total & 212,559 & 2,850 \\
\bottomrule
\end{tabular}
\end{table}

\begin{table*}[t]
\centering
\caption{Canonical route-to-specialist mapping and held-out evaluation inventory. The held-out set contains 190 questions per route except BNG/subscriber management with 380.}
\label{tab:routes}
\scriptsize
\setlength{\tabcolsep}{3pt}
\begin{tabular}{llrr@{\hspace{0.7em}}llrr}
\toprule
Specialist & Route & Train & Eval. & Specialist & Route & Train & Eval. \\
\midrule
Dynamic Routing & BGP & 19,756 & 190 & CLI Diagnostics & Linux tool calling & 5,817 & 190 \\
Dynamic Routing & OSPF & 9,561 & 190 & CLI Diagnostics & Linux agentic diagnosis & 5,817 & 190 \\
Dynamic Routing & IS-IS & 16,280 & 190 & CLI Diagnostics & Network fundamentals & 9,356 & 190 \\
MPLS Services & MPLS VPN & 26,439 & 190 & Edge Access & L2 switching & 11,061 & 190 \\
Broadband Services & BNG/subscriber mgmt. & 24,735 & 380 & Edge Access & PON/optical & 8,296 & 190 \\
Broadband Services & QoS & 4,625 & 190 & Edge Access & WiFi gateway & 18,733 & 190 \\
Routing Fundamentals & IP routing & 23,484 & 190 & Routing Fundamentals & Multicast & 28,599 & 190 \\
\bottomrule
\end{tabular}
\end{table*}

\subsection{Shared CPT Backbone and Specialist LoRAs}
The shared backbone starts from a pinned Meta-Llama-3.1-8B-Instruct snapshot and undergoes full-parameter continued pretraining over the fourteen routed training files. The executed CPT run consumes 212,559 records, records 47,975,711 training tokens, and completes 3,322 optimizer steps. The CPT stage uses sequence length 1,024, one epoch, bfloat16 precision, fused AdamW, learning rate $10^{-5}$ with a linear schedule and 10 warmup steps, per-device batch size 1, gradient accumulation 16, and an effective global batch size of 64 across four GPUs.

Each of the six specialist teachers is then trained as a separate LoRA adapter over the same CPT checkpoint. Each specialist consumes only the routed training records assigned to its specialist group; across all six specialists these assigned subsets partition the same 212,559-record routed training inventory used by CPT, while the SFT stage applies instruction formatting and three specialist-specific epochs rather than another full-corpus CPT pass. All specialists use rank 64, $\alpha=128$, dropout 0, learning rate $2\times10^{-5}$, weight decay 0, maximum gradient norm 1.0, micro-batch size 1, gradient accumulation 8, and three SFT epochs with seed 42. The adapted modules are \texttt{q\_proj}, \texttt{k\_proj}, \texttt{v\_proj}, \texttt{o\_proj}, \texttt{gate\_proj}, \texttt{up\_proj}, and \texttt{down\_proj}; maximum sequence length ranges from 1,024 to 1,536 according to specialist domain.

During RSOD, teacher adapters, token embeddings, backbone, final normalization, and output head remain frozen; only the student LoRA parameters receive gradients. Compatibility therefore requires a common base identity, tokenizer and vocabulary ordering, final normalization, and LM-head/output-projection geometry; these are method requirements for the current RSOD reconstruction path. For an adapted projection $W_0$, specialist $k$ and the student use separate low-rank updates:
\begin{equation}
W_k = W_0 + s_k B_k A_k,
\qquad
W_s = W_0 + s_s B_s A_s.
\label{eq:lora}
\end{equation}
RSOD performs behavioral consolidation through distillation while the specialist adapters remain frozen and separate.

\section{Routed Specialist On-Policy Distillation}
\textbf{RSOD overview.} Routed Specialist On-Policy Distillation (RSOD) consolidates multiple frozen domain-specialist LoRA teachers into one student LoRA without merging teacher weights. The current student generates its own trajectory, a deterministic route map selects exactly one specialist teacher, and that teacher scores the student's exact native token sequence by teacher forcing. The teacher returns final hidden states rather than dense vocabulary logits; the trainer reconstructs the teacher distribution locally through the shared frozen final normalization and LM head, computes forward KL on generated positions, updates only the student LoRA, and refreshes the new student version before the next rollout.

\subsection{Routed Student-On-Policy Supervision}
At optimizer step $n$, the deterministic scheduler selects route $r$ and prompt $x$. A fixed ownership map $\pi(r)$ selects specialist $k=\pi(r)$. The committed student version $v_n$ generates trajectory $y=(y_1,\ldots,y_T)$, and teacher $T_{\pi(r)}$ teacher-forces the complete native token sequence $(x,y)$. The teacher therefore scores exactly the tokens produced by the student. Prompt tokens provide context and generated positions contribute to distillation.

The trajectory is produced by the current student and evaluated by the assigned specialist on the student's exact tokenization. Route scheduling is used during training to select supervision; deployment uses the single consolidated student adapter.

\begin{figure*}[t]
    \centering
    \includegraphics[width=0.99\textwidth]{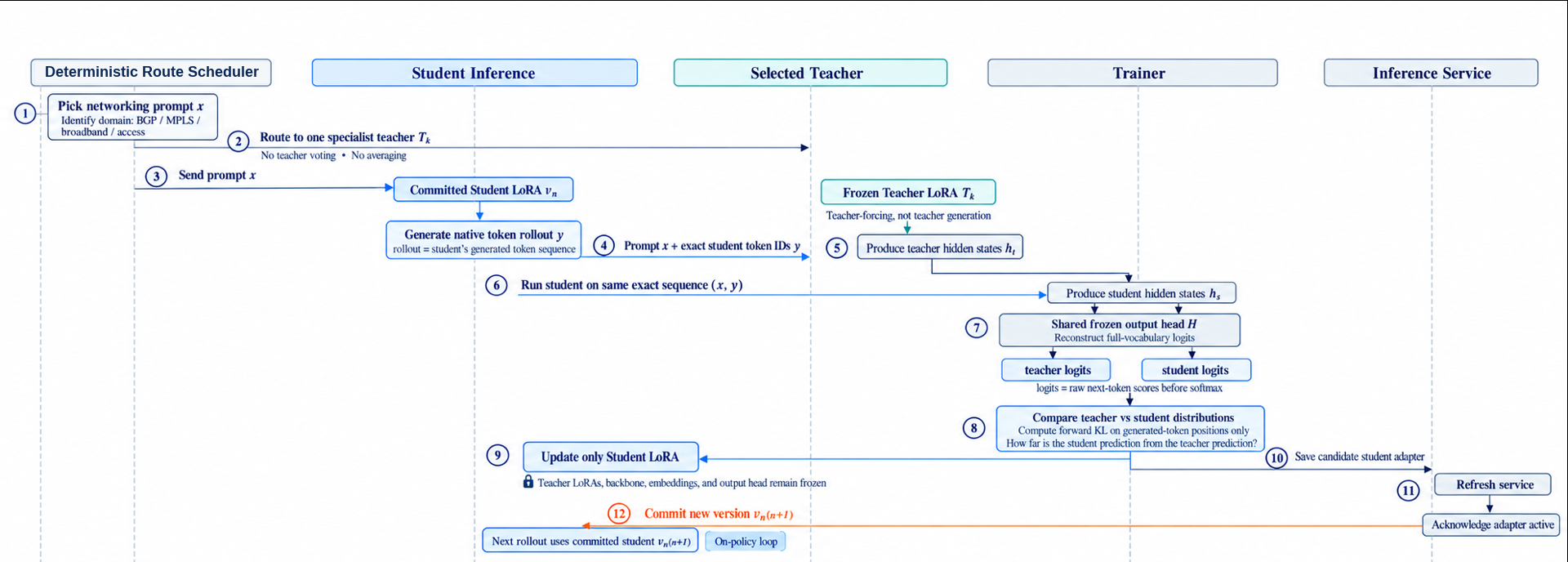}
    \caption{Detailed RSOD execution sequence. The deterministic route scheduler chooses one frozen specialist teacher; the committed student generates the native rollout, and the teacher scores the same token sequence by teacher forcing. Teacher and student hidden states are mapped through the shared frozen normalization/output head, forward KL is computed on generated positions, and only the student LoRA is updated. The candidate adapter is refreshed, acknowledged active, and committed before the next rollout.}
    \label{fig:mopd}
\end{figure*}

\subsection{Compact Teacher Signal and KL Objective}
The inference service returns final teacher hidden states $h_T$ instead of a dense vocabulary-logit tensor. Let $g(\cdot)$ be the model's frozen final normalization and $W_{\mathrm{LM}}$ the shared frozen LM-head/output projection. The trainer reconstructs distributions locally:
\begin{align}
p_T &= \operatorname{softmax}\!\left(\frac{W_{\mathrm{LM}}\,g(h_T)}{\tau}\right), \\
p_S &= \operatorname{softmax}\!\left(\frac{W_{\mathrm{LM}}\,g(h_S)}{\tau}\right).
\end{align}
For $N$ trajectories with generated lengths $T_i$, the evaluated configuration uses forward KL:
\begin{equation}
\begin{aligned}
\mathcal{L}
&= \frac{\lambda\tau^2}{N}
\sum_{i=1}^{N}\frac{1}{T_i}
\sum_{t=1}^{T_i} \\
&\quad D_{\mathrm{KL}}\!\left(
 p_T(\cdot\mid x_i,y_{i,<t})
 \;\Vert\;
 p_S(\cdot\mid x_i,y_{i,<t})
\right).
\end{aligned}
\label{eq:kl}
\end{equation}
This is logit-level KD reconstructed from hidden states rather than hidden-state matching. Forward KL preserves the teacher's probability mass over the student's own trajectory, providing richer supervision than a hard response target while retaining the on-policy training distribution \cite{agarwal2023gkd}. The teacher signal transported per position has hidden width $d$ rather than vocabulary width $|V|$. Across the run, the trainer materialized 67.86~GB of FP32 teacher hidden-state tensors for full sequences; the analytical dense FP32 logit quantity over the 2,926,443 generated/KD positions is 1.50~TB. Figure~\ref{fig:systems} reports these as complementary teacher-signal accounting views.

\subsection{Compact-Signal Fidelity Validation}
We validate the hidden-state transport path independently with a controlled single-GPU RSOD test that executes fresh student initialization, student rollout, teacher forcing, shared LM-head logit reconstruction, KL computation, backpropagation, and a student update. For a 50-token trajectory, native teacher logits and logits reconstructed as $W_{\mathrm{LM}}\,g(h_T)$ are compared at all 50 generated positions over the full 128,256-token vocabulary. The reconstructed logits show mean absolute error $7.26\times10^{-6}$, median absolute error 0, maximum absolute error $6.25\times10^{-2}$, cosine similarity 1.000000, and 100\% top-1 agreement. Exact top-5 and top-10 ranking matches are 98\% and 94\%, respectively. The native-versus-reconstructed distribution KL is numerically zero to floating-point precision ($-4.04\times10^{-9}$ as computed), while the student-versus-teacher KD loss remains nontrivial at 1.1992. Student parameters receive gradients and change after the optimizer step. These measurements show that shared-head reconstruction preserves the teacher prediction distribution to high numerical fidelity while retaining a usable distillation signal for the student.

\subsection{Version-Consistent Adapter Commit}
After optimization, the student adapter is saved as a candidate, loaded by the tensor-parallel inference service, acknowledged with a changed service identity, and atomically committed before the next rollout. This enforces a version-consistency invariant: the next trajectory must come from the acknowledged committed student. Two additional invariants complete the execution contract: teacher forcing consumes the exact native student token IDs, and transported teacher hidden states reconstruct the intended teacher distribution through the shared frozen output components. The run records expected and serving adapter versions, hashes student-generated and teacher-forced token arrays, and retains route, configuration, synchronization, and telemetry evidence. All 2,000 commits completed successfully and all 16,000 recorded version and token-identity checks matched; the controlled reconstruction test independently validates teacher-signal fidelity. Mean prepare-to-commit latency is 6.47~s versus 7.65~s for student rollout.

\section{Memory-Conscious Execution}
\subsection{Execution Topology}
The evaluated workstation contains four NVIDIA RTX 6000 Ada Generation GPUs (49,140~MiB each). Two GPUs run DDP student optimization; the other two host a TP2 inference service containing one shared backbone, six frozen teacher adapters, and the current student adapter. The system is evaluated as a workstation-scale four-GPU configuration.

For $K$ compatible teachers and one student adapter, shared inference weight storage is approximately
\begin{equation}
M_{\mathrm{shared}} \approx M_B + \sum_{k=1}^{K} M_{A_k} + M_{A_s}.
\end{equation}
A deliberately replicated-teacher reference would require approximately
\begin{equation}
M_{\mathrm{replicated}} \approx K M_B + \sum_{k=1}^{K} M_{A_k} + M_{A_s}.
\end{equation}
The implementation uses resident specialist adapters over one shared inference backbone. The central memory property is therefore \emph{six specialists, not six backbones}: the TP2 inference pair keeps one 8B backbone resident and overlays six frozen teacher LoRAs plus the current student LoRA. At BF16, an 8B parameter payload is roughly 16~GB before runtime overheads, so avoiding one full backbone per teacher removes the dominant $K M_B$ replication term. On the DDP training pair, each rank holds a frozen backbone replica and trains only the rank-64 student LoRA; backbone gradients and backbone optimizer state are not maintained. The four-GPU design therefore combines shared teacher residency, LoRA-only student optimization, TP inference, DDP training, and compact hidden-state transfer rather than dense teacher-logit transfer. Measured peaks were approximately 34.3 and 34.0~GiB on the two DDP GPUs and 47.5~GiB on each TP inference GPU, for 159.5~GiB aggregate peak usage. The equations describe weight residency; measured VRAM includes activations, KV cache, allocator reserve, runtime workspaces, and adapter-serving overhead. We therefore characterize the architecture as \emph{memory-conscious}.

\subsection{Measurement Methodology}
GPU memory, utilization, and power are sampled through NVML. Energy is estimated by trapezoidal integration over valid power samples; intervals larger than 2.5 times the per-GPU median sampling interval are counted as telemetry gaps and excluded rather than silently interpolated. Phase latencies are rank-0 host-boundary observations without inserted CUDA synchronization; they characterize the end-to-end control path rather than individual GPU kernels.

Raw telemetry, step timings, trajectory metrics, adapter events, version-commit events, token/version audits, checkpoint metrics, and environment/model/data manifests are retained as the source of truth. Exact artifact hashes are intentionally kept out of the paper body and belong in the release manifest.

\section{Networking Evaluation}
\subsection{Evaluation Conditions}
The primary evaluation is route-conditioned by design: each held-out item is paired with a fixed route-specific system prompt corresponding to its canonical networking route. The same item-level route prompt is used for every compared model, so raw-model, CPT+specialist-teacher, and student scores are evaluated under the same held-out manifest, prompt construction, decoding setup, judge configuration, aggregation, and failure policy. This measures answer quality under the intended operating condition in which route-specific expertise is supplied to the model. We additionally evaluate the final student under a neutral deployment-like prompt, using the same benchmark and scoring framework, to quantify performance without the route-specific expert instruction.

The 2,850-question held-out inventory is stored separately from the routed training files. Each route contributes 190 questions except BNG/subscriber management, which contributes 380. The judge configuration uses the Anthropic alias \texttt{claude-sonnet-4-5} with temperature 0, a 2,000-token judge response limit, and a 60-second timeout.

\subsection{Route-Aware Judge and Validation Policy}
Evaluation uses a route-aware five-dimension rubric. Each judge call receives the route/category, question, and generated response together with an ``expert network engineer'' instruction and evaluates up to five anonymized responses per batch. Model, adapter, teacher, and student identities are not included in the judge prompt. The required JSON output contains numeric scores from 0 to 5 for factual correctness, completeness, terminology/command precision, hallucination control, and operational usefulness; the scorer derives and validates the 0--25 total.

The hallucination dimension is reserved for technically false or fabricated claims, while missing or vague content is scored under completeness or terminology. This separation reduces double-penalization and keeps the rubric aligned with operational network-engineering quality. Malformed or inconsistent outputs are rejected under a bounded policy of up to three total attempts, with one- and two-second waits between attempts. Validation rejects wrong item count/order, malformed JSON, missing or out-of-range dimensions, inconsistent totals, and five-item all-zero batches. The scorer records coverage separately from quality so rejected judge batches do not silently become zero-valued model scores.

\subsection{Aggregation}
For model $m$, the micro average is
\begin{equation}
\operatorname{micro}(m)=\frac{1}{|I_m|}\sum_{i\in I_m}s_{mi},
\end{equation}
and the route macro average is
\begin{equation}
\operatorname{macro}(m)=\frac{1}{|R|}\sum_{r\in R}\bar{s}_{mr}.
\end{equation}
Coverage and judge failures are reported separately from score. The matched primary comparison covers the raw model, the six routed CPT+specialist teachers, and the consolidated RSOD student over the same 2,850-question manifest; the student is also reported under the separate neutral prompting condition.

\section{Experimental Setup}
The RSOD run uses the shared 8B CPT backbone artifact \texttt{cpt-llama31-8b-six-expert-final-oct-1} and a fresh student LoRA with rank 64, $\alpha=128$, dropout 0, and the same seven attention/MLP projection targets as the specialists; \texttt{lm\_head} and \texttt{embed\_tokens} are protected from adaptation. The deterministic fourteen-route schedule assigns each trajectory to exactly one of the six frozen specialist teachers.

The distillation objective is forward KL with temperature $\tau=1.0$, weight $\lambda=1.0$, and generated-token-only masking. Student optimization uses AdamW with learning rate $10^{-5}$, cosine scheduling, 3\% warmup, weight decay 0.01, gradient clipping at 1.0, and seed 42. Rollouts use temperature 0, top-$p=1.0$, a maximum of 256 generated tokens, and one trajectory per prompt. Eight trajectories are processed per optimizer step for 2,000 optimizer steps and 16,000 trajectories total; checkpoints are recorded every 250 steps.

The four-GPU placement dedicates GPUs 0--1 to DDP student optimization and GPUs 2--3 to TP2 inference. The inference service can keep the six teachers and current student LoRA resident over the shared backbone. Runtime configuration, hardware/software identities, model identities, dataset manifests, telemetry, checkpoint metadata, judge rubrics, and evaluation reports are retained as machine-readable run artifacts.

\begin{figure*}[t]
    \centering
    \includegraphics[width=0.98\textwidth]{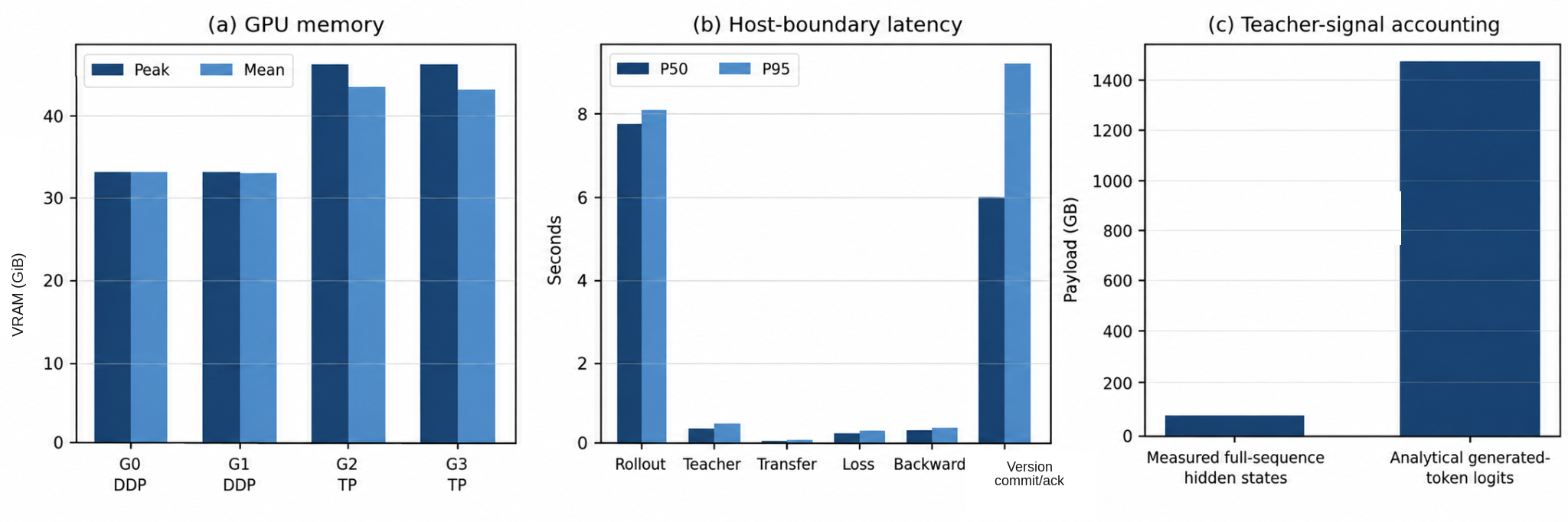}
    \caption{Systems summary. (a) Mean and peak device VRAM across the four GPUs. (b) P50/P95 rank-0 host-boundary latency across major phases, including version commit/acknowledgement. (c) Trainer-side hidden-state materialization and analytical dense FP32 logit accounting.}
    \label{fig:systems}
\end{figure*}

\begin{figure*}[t]
    \centering
    \includegraphics[width=0.98\textwidth]{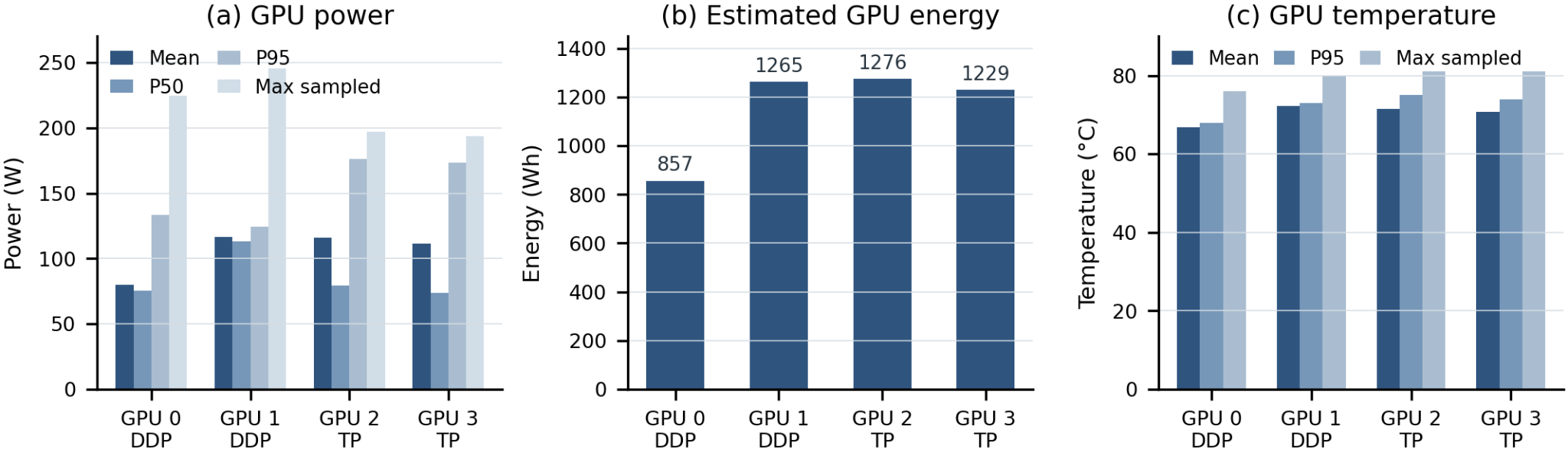}
    \caption{Per-device telemetry for the run. (a) GPU power statistics. (b) Estimated GPU energy by device from NVML sampling. (c) GPU temperature statistics.}
    \label{fig:telemetry}
\end{figure*}

\section{Results}
\subsection{Systems Results}
Table~\ref{tab:systems} summarizes the run. It consumed 56.1 GPU-hours and an NVML-derived 4.63~kWh estimate. Peak aggregate device memory was 159.5~GiB across the four GPUs. All 2,000 adapter commits completed successfully, and all 16,000 recorded student-version and token-identity checks matched.

\begin{table}[t]
\centering
\caption{RSOD run summary. Energy is NVML-derived; VRAM is aggregate device usage.}
\label{tab:systems}
\small
\begin{tabular}{lr}
\toprule
Metric & Value \\
\midrule
Optimizer steps & 2,000 \\
Trajectories & 16,000 \\
KD tokens & 2,926,443 \\
Active training time & 50,511 s \\
GPU-hours & 56.1 \\
Estimated GPU energy & 4.63 kWh \\
Peak aggregate VRAM & 159.5 GiB \\
Trajectories/s & 0.317 \\
KD tokens/s & 57.9 \\
Adapter commits & 2,000 / 2,000 \\
Version mismatches & 0 / 16,000 \\
Token-ID mismatches & 0 / 16,000 \\
\bottomrule
\end{tabular}
\end{table}

Figure~\ref{fig:systems} condenses the main systems evidence. Teacher inference and signal transfer are small compared with rollout and version-commit/acknowledgement latency. Teacher-signal accounting records 67.86~GB of trainer-materialized FP32 hidden-state tensors, while the corresponding analytical dense-logit accounting is 1.50~TB. The figure reports these trainer-side accounting quantities alongside the measured execution data.

The full run averaged 25.3~s per optimizer step (50,511~s over 2,000 steps). The instrumented step phases report a lower mean of 22.36~s because they exclude run-level overhead outside the timed step sections. Within the instrumented phases, student rollout and version commit/acknowledgement are the largest measured components of the train--serve loop.

Figure~\ref{fig:telemetry} adds per-device power, energy, and temperature telemetry. The two tensor-parallel inference GPUs and one of the DDP GPUs account for the largest energy totals, while GPU~0 shows lower average power and energy than the other three devices. Temperatures remain within the sampled operating range shown in the figure.

\begin{figure*}[!t]
    \centering
    \includegraphics[width=0.98\textwidth]{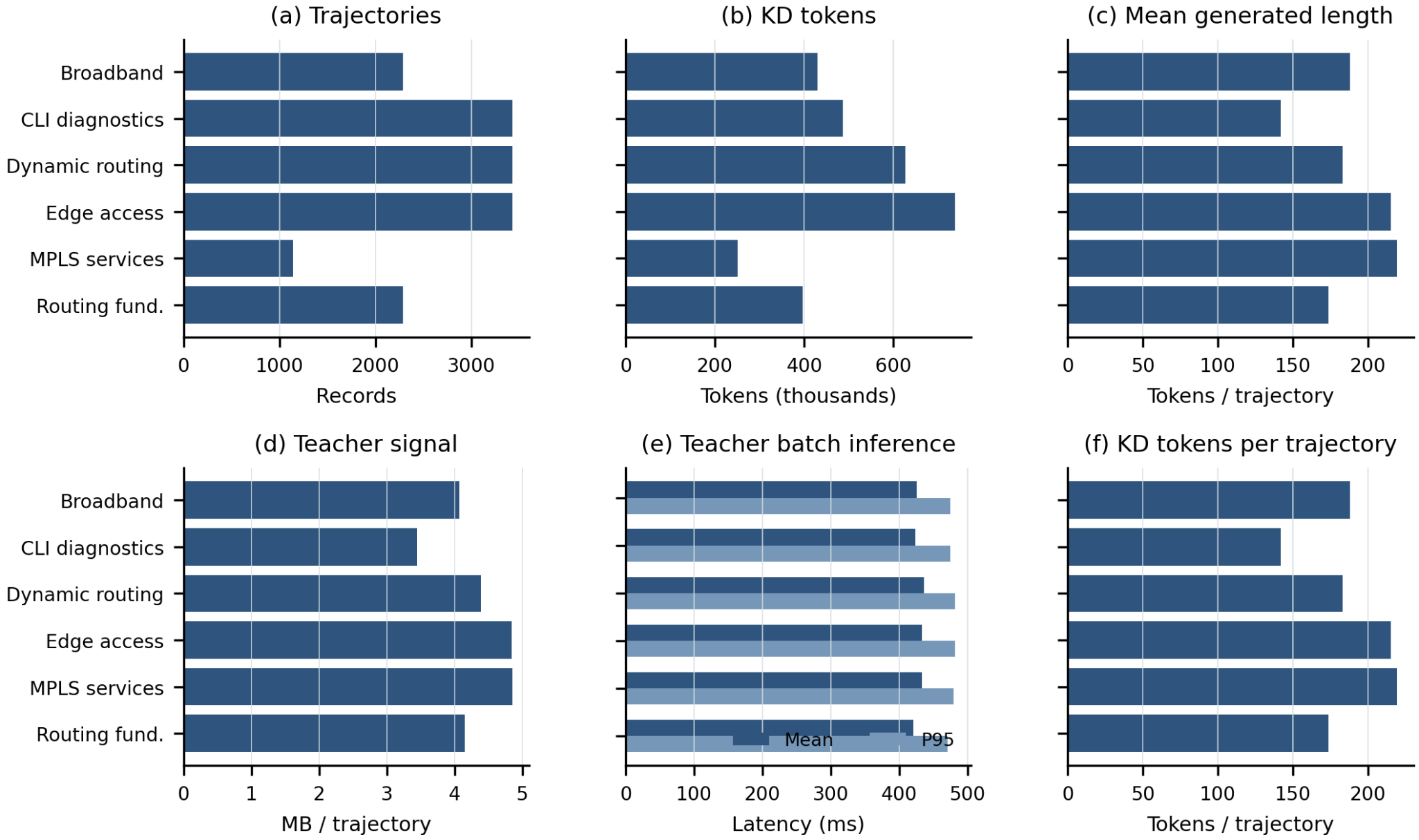}
    \caption{Per-specialist-group workload and runtime statistics. Panels show trajectory counts, KD-token totals, mean generated length, mean teacher-signal size, teacher batch-inference latency, and mean KD tokens per trajectory.}
    \label{fig:route_runtime}
\end{figure*}

Figure~\ref{fig:route_runtime} breaks the runtime down by specialist group. CLI diagnostics, dynamic routing, and edge access receive the largest trajectory counts, while edge access and dynamic routing account for the largest KD-token and teacher-signal volumes. Teacher batch-inference latency remains relatively tight across groups, despite these workload differences. This view helps explain which specialist groups dominate traffic and transport volume during the run.

\begin{figure*}[!t]
    \centering
    \includegraphics[width=0.92\textwidth]{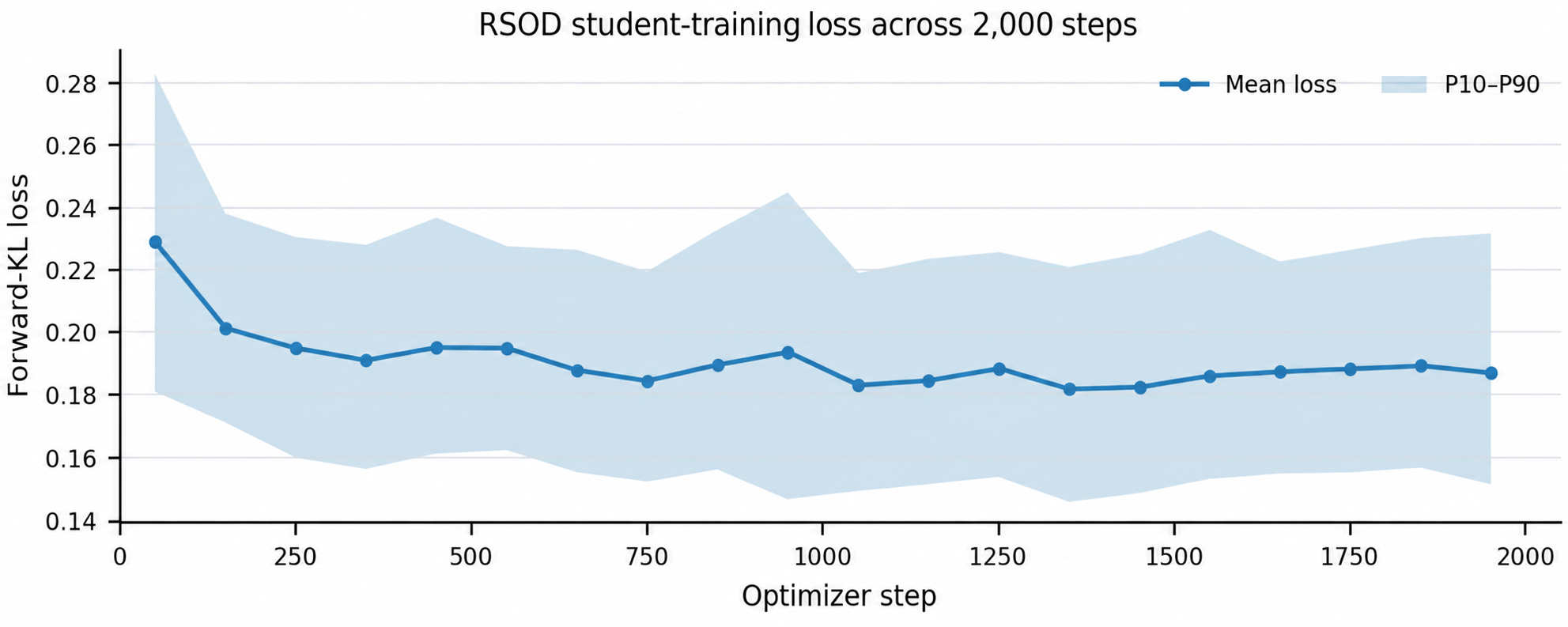}\\[-0.2em]
    \begin{minipage}[t]{0.48\textwidth}
        \centering
        \includegraphics[width=\textwidth]{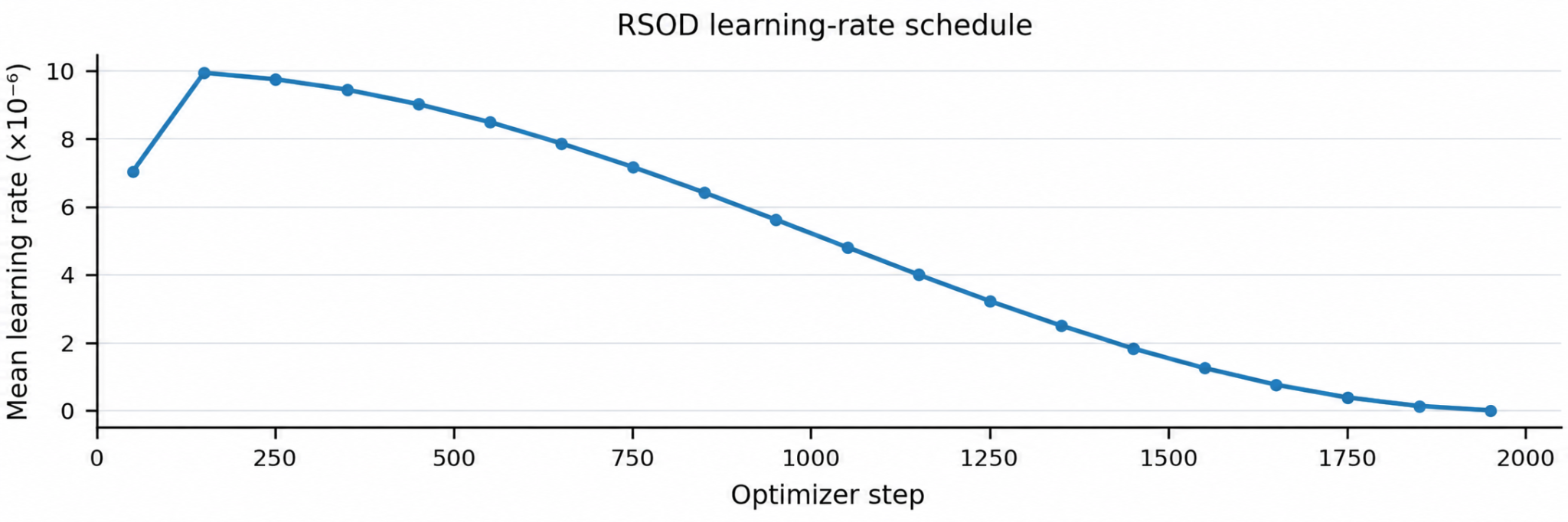}
    \end{minipage}\hfill
    \begin{minipage}[t]{0.48\textwidth}
        \centering
        \includegraphics[width=\textwidth]{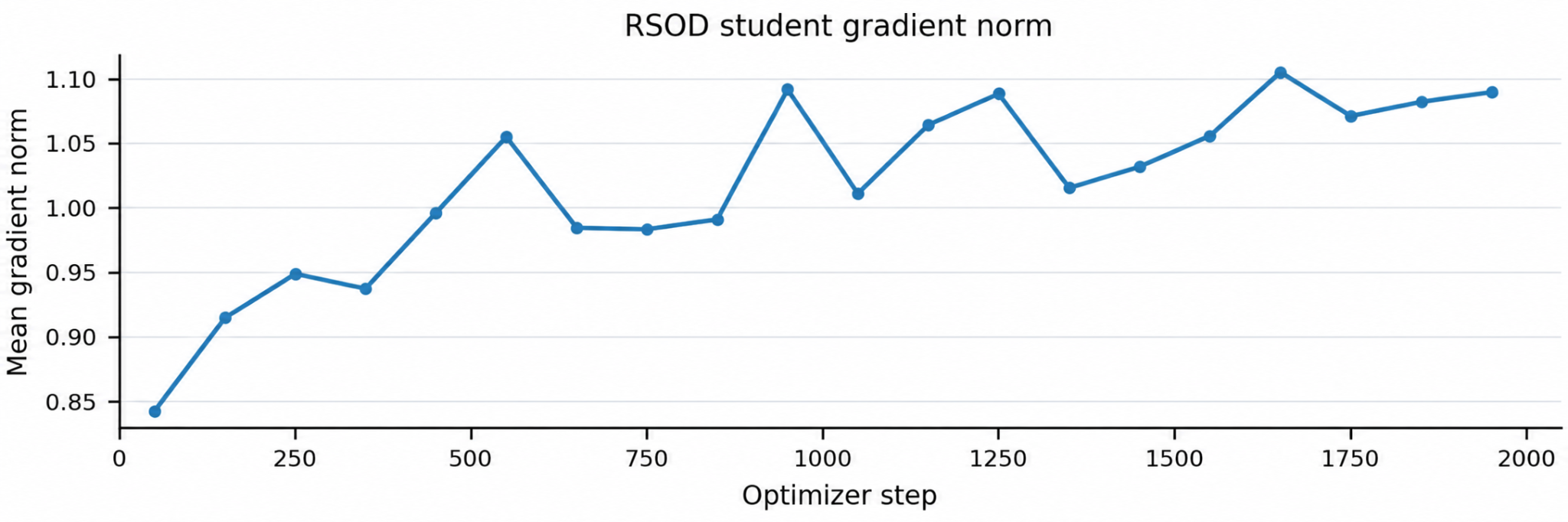}
    \end{minipage}
    \caption{Optimization traces over 2,000 steps: forward-KL training loss with P10--P90 band (top), learning-rate schedule (bottom left), and mean student gradient norm (bottom right).}
    \label{fig:training_dynamics}
\end{figure*}

Figure~\ref{fig:training_dynamics} shows the optimization traces. Forward-KL loss drops early and stabilizes around the high-0.18 range, the learning-rate schedule follows warmup and smooth decay, and the mean student gradient norm remains in a narrow operating band through the final step.
\FloatBarrier

\raggedbottom
\subsection{Model Quality Across Training Stages}
The primary route-conditioned evaluation uses the same frozen 2,850-question manifest and the same item-level route-specific system prompts for the raw Meta-Llama-3.1-8B-Instruct model, the six CPT+specialist teachers, and the consolidated RSOD student. Generation uses deterministic decoding (temperature 0, top-$p$ 1.0, no sampling) with a maximum of 512 newly generated tokens. The raw model scores 12.8661/25 micro and 13.0172/25 macro. When each question is answered by its assigned specialist, the combined routed six-teacher system scores 21.7641/25 micro over 2,850/2,850 questions; the six specialist-group means span 19.0368--22.5912/25 with complete group coverage. The final RSOD student reaches 21.7112/25 micro and 21.8006/25 macro, an 8.8451-point micro-score gain over the raw model and a 0.0529-point micro difference from the combined routed-teacher system. Table~\ref{tab:quality_groups} reports the detailed specialist-group results.

\enlargethispage{1.5in}
\vspace{0.65em}
\noindent\begin{minipage}{\columnwidth}
\centering
\captionsetup{font=footnotesize,skip=1pt}
\captionof{table}{Quality scores for the raw model, routed specialists, and Llama-CalixCLM v1.}
\label{tab:quality_groups}
\fontsize{6.4}{6.9}\selectfont
\renewcommand{\arraystretch}{0.82}
\setlength{\tabcolsep}{2.3pt}
\begin{tabular}{lrrr}
\toprule
Model / specialist group & Micro/group mean & Macro & Scored/Total \\
\midrule
Raw model & 12.8661 & 13.0172 & 2845/2850 \\
\textbf{Combined routed teachers} & \textbf{21.7641} & -- & \textbf{2850/2850} \\
Dynamic Routing teacher & 22.5912 & -- & 570/570 \\
Edge Access teacher & 22.5316 & -- & 570/570 \\
CLI Diagnostics teacher & 21.8544 & -- & 570/570 \\
Broadband Services teacher & 21.5276 & -- & 570/570 \\
Routing Fundamentals teacher & 20.9553 & -- & 380/380 \\
MPLS Services teacher & 19.0368 & -- & 190/190 \\
Llama-CalixCLM v1 (route) & 21.7112 & 21.8006 & 2850/2850 \\
Llama-CalixCLM v1 (neutral) & 20.0028 & 20.0778 & 2850/2850 \\
\bottomrule
\end{tabular}
\end{minipage}
\vspace{0.65em}
\newpage

A secondary neutral deployment-like prompting condition evaluates the same final student without the route-specific expert instruction. Under this condition the student reaches 20.0028/25 micro and 20.0778/25 macro. This secondary result reports deployment-like behavior without the route-specific specialization cue; no raw-model neutral baseline is used for a relative claim. Table~\ref{tab:quality_groups} and Figure~\ref{fig:quality} summarize the primary progression and the secondary student condition.

\begin{center}
    \includegraphics[width=0.98\columnwidth]{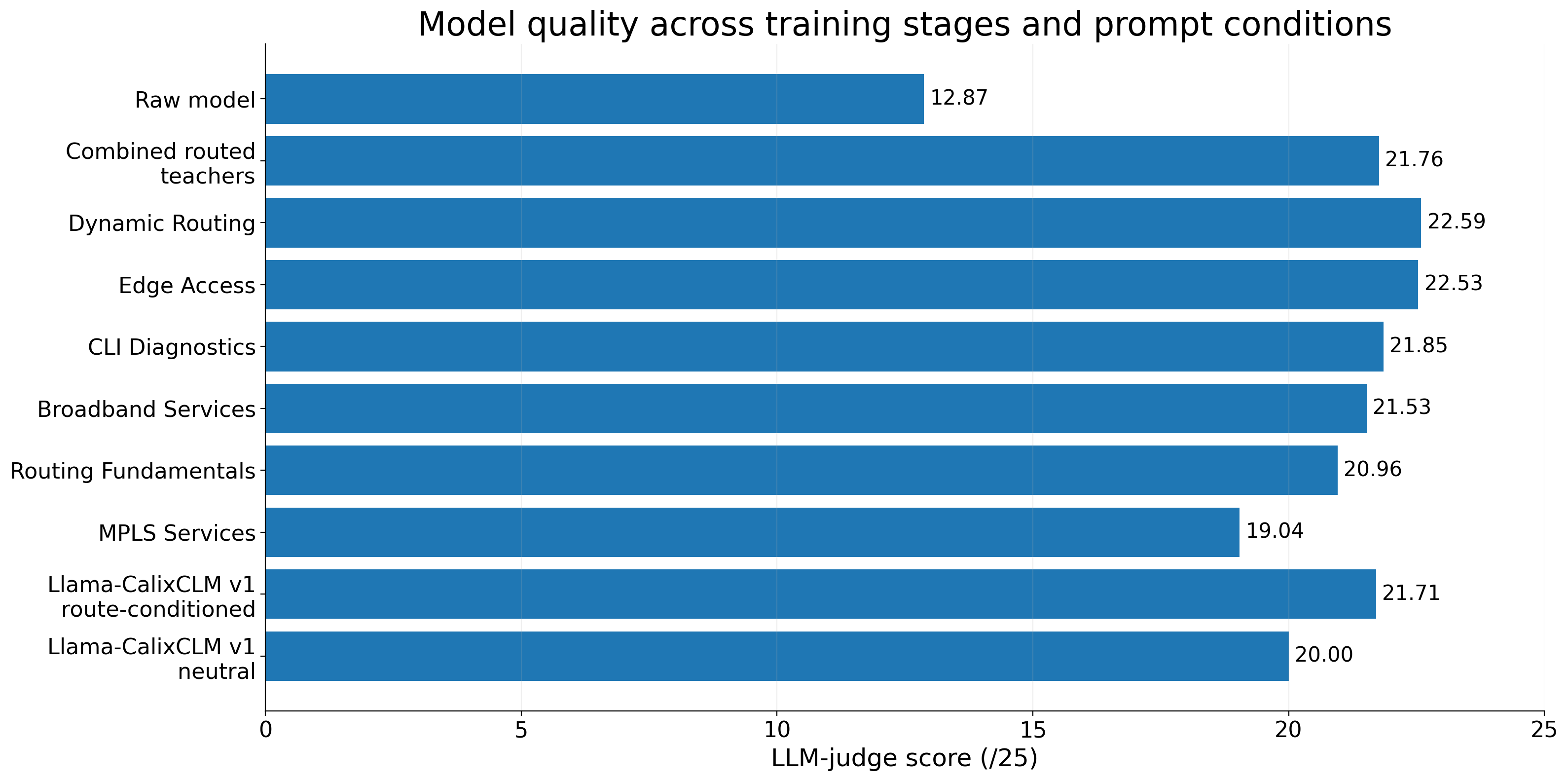}\par
    \captionof{figure}{LLM-judge scores for the raw model, combined routed teacher system, specialist groups, and consolidated RSOD student. Route-conditioned prompting is the primary comparison; the neutral student bar is a secondary deployment-like condition.}
    \label{fig:quality}
\end{center}
\flushbottom

\section{Reproducibility}
Machine-readable manifests link the raw base, CPT backbone, routed datasets, specialist adapters, resolved RSOD configuration, telemetry, checkpoints, judge rubrics, and evaluation reports. The executed CPT record fixes the fourteen inputs, tokenizer identity, 212,559 training records, 47,975,711 consumed tokens, and 3,322 optimizer steps; dataset manifests preserve per-route counts and SHA-256 identities for train and held-out artifacts. The resolved RSOD configuration fixes route ownership, optimizer/KD/rollout settings, GPU placement, and student LoRA geometry. Together with version checks, token hashes, synchronization events, and reconstruction validation, these artifacts provide the evidence layer for the model-version, token-identity, and teacher-signal invariants.

\paragraph{Released artifacts.} Calix AI Labs publishes model artifacts through the Hugging Face organization at \url{https://huggingface.co/calix-ai-labs}. The TelcoBench evaluation framework is available at \url{https://github.com/calix-ai-labs/telcobench}.

\section{Limitations}
Response quality is measured with an LLM-judge rubric, so the reported 0--25 values are rubric-based model-quality scores rather than executable correctness guarantees. The paper reports both the primary route-conditioned protocol and a secondary neutral deployment-like prompting condition for the final student.

\section{Conclusion}
We present an end-to-end telecom language-model workflow spanning source-grounded structured data generation, shared continued pretraining, specialist LoRA post-training, routed specialist on-policy distillation, and networking evaluation. The executed data pipeline materializes 212,559 routed training records across fourteen routes and six specialists. RSOD consolidates the six frozen specialists into one deployable student adapter: the student generates the trajectory, the deterministic route scheduler selects one specialist, the teacher scores the exact native token sequence, and only the student LoRA is optimized. A controlled reconstruction test validates the compact teacher-signal path with $7.26\times10^{-6}$ mean absolute logit error and 100\% top-1 agreement between native and reconstructed teacher logits. Across 16,000 training trajectories, all recorded token-identity and serving-version checks match. Under the primary route-conditioned evaluation, the raw model scores 12.8661/25 micro, the combined routed six-teacher system scores 21.7641/25 micro over all 2,850 questions, and Llama-CalixCLM v1 reaches 21.7112/25 micro and 21.8006/25 macro; individual specialist-group means span 19.0368--22.5912/25. Under the separate neutral deployment-like prompt, the same student reaches 20.0028/25 micro and 20.0778/25 macro. Together, the data, numerical-fidelity, quality, audit, and systems measurements provide a practical path from domain-specialized networking knowledge to a single telecom-focused deployment adapter.

\FloatBarrier

\onecolumn
\appendix
\section{Reproducibility Summary}
\begin{center}
\small
\begin{tabular}{p{0.22\textwidth}p{0.70\textwidth}}
\toprule
Item & Recorded value / boundary \\
\midrule
Backbone artifact & \texttt{cpt-llama31-8b-six-expert-final-oct-1}; CPT: 212,559 routed records, 47,975,711 tokens, 3,322 steps \\
Specialists / student & Six frozen LoRA teachers and one fresh student; rank 64, alpha 128, seven attention/MLP projection targets \\
Hardware & 4$\times$ NVIDIA RTX 6000 Ada; GPUs 0--1 DDP student, GPUs 2--3 TP inference \\
RSOD recipe & $\tau=1$, KL weight 1, generated-token-only mask; AdamW $10^{-5}$, cosine schedule, 3\% warmup; 2,000 steps, 16,000 trajectories \\
Version commit & candidate save $\rightarrow$ service refresh $\rightarrow$ acknowledgement $\rightarrow$ COMMITTED $\rightarrow$ next rollout \\
Audit & 2,000/2,000 adapter commits completed; 16,000 version checks and 16,000 token-ID checks, zero observed mismatches \\
Compact-signal fidelity & 50 generated positions, 128,256-token vocabulary; MAE $7.26\times10^{-6}$; 100\% top-1 agreement; 98\% exact top-5 match \\
Evaluation generation & Route-specific expertise prompt; native chat template; deterministic generation \\
Judge & \texttt{claude-sonnet-4-5}; temperature 0; batch size 5; five 0--5 rubric dimensions; identity-masked prompt; validated JSON; up to three total attempts \\
Evaluation manifest & 2,850 questions; 190 per route except BNG/subscriber management with 380 \\
\bottomrule
\end{tabular}
\end{center}

\end{document}